\documentclass[10pt,twocolumn,letterpaper]{article}

\usepackage{cvpr}              

\usepackage{multirow}
\usepackage{graphicx}      
\usepackage{booktabs}      
\usepackage{multirow}      
\usepackage{makecell}      
\usepackage{array}  
\usepackage[table]{xcolor} 
\usepackage{pifont}

\definecolor{cvprblue}{rgb}{0.21,0.49,0.74}
\usepackage[pagebackref,breaklinks,colorlinks,allcolors=cvprblue]{hyperref}
\usepackage{bm}

\usepackage{soul}
\usepackage{xcolor}

\def\confName{CVPR}
\def\confYear{2026}

\title{Face-Guided Sentiment Boundary Enhancement for Weakly-Supervised Temporal Sentiment Localization}
\author{
Cailing Han$^{1}$\thanks{Equal contribution.} \quad
Zhangbin Li$^{1}$\footnotemark[1] \quad
Jinxing Zhou$^{2}$\thanks{Corresponding author.} \\
Wei Qian$^{1}$ \quad
Jingjing Hu$^{1}$ \quad
Yanghao Zhou$^{3}$ \quad
Zhangling Duan$^{4}$ \quad
Dan Guo$^{1,4,5}$\footnotemark[2]\\
$^{1}$Hefei University of Technology \quad
$^{2}$MBZUAI\quad
$^{3}$NUS \\
$^{4}$Institute of Artificial Intelligence, Hefei Comprehensive National Science Center\\
$^{5}$The Key Laboratory of Knowledge Engineering with Big Data, Hefei University of Technology\\
{\tt \{ceilinghan, lizhangbin.mail, zhoujxhfut\}@gmail.com, guodan@hfut.edu.cn}  
}
\begin{document}
\maketitle
\begin{abstract}
Point-level weakly-supervised temporal sentiment localization (P-WTSL) aims to detect sentiment-relevant segments in untrimmed multimodal videos using timestamp sentiment annotations, which greatly reduces the costly frame-level labeling.
To further tackle the challenges of imprecise sentiment boundaries in P-WTSL, we propose the Face-guided Sentiment Boundary Enhancement Network (\textbf{FSENet}), a unified framework that leverages fine-grained facial features to guide sentiment localization.
Specifically, our approach \textit{first} introduces the Face-guided Sentiment Discovery (FSD) module, which integrates facial features into multimodal interaction via dual-branch modeling for effective sentiment stimuli clues;
We \textit{then} propose the Point-aware Sentiment Semantics Contrast (PSSC) strategy to discriminate sentiment semantics of candidate points (frame-level) near annotation points via contrastive learning, thereby enhancing the model's ability to recognize sentiment boundaries.
At \textit{last}, we design the Boundary-aware Sentiment Pseudo-label Generation (BSPG) approach to convert sparse point annotations into temporally smooth supervisory pseudo-labels.
Extensive experiments and visualizations on the benchmark demonstrate the effectiveness of our framework, achieving state-of-the-art performance under full supervision, video-level, and point-level weak supervision, thereby showcasing the strong generalization ability of our FSENet across different annotation settings. Code: \url{https://github.com/CeilingHan/FSENet}.
\end{abstract}    
\section{Introduction}
\label{sec:intro}
\begin{figure}[!t]
\centering
\includegraphics[width=0.99\linewidth]{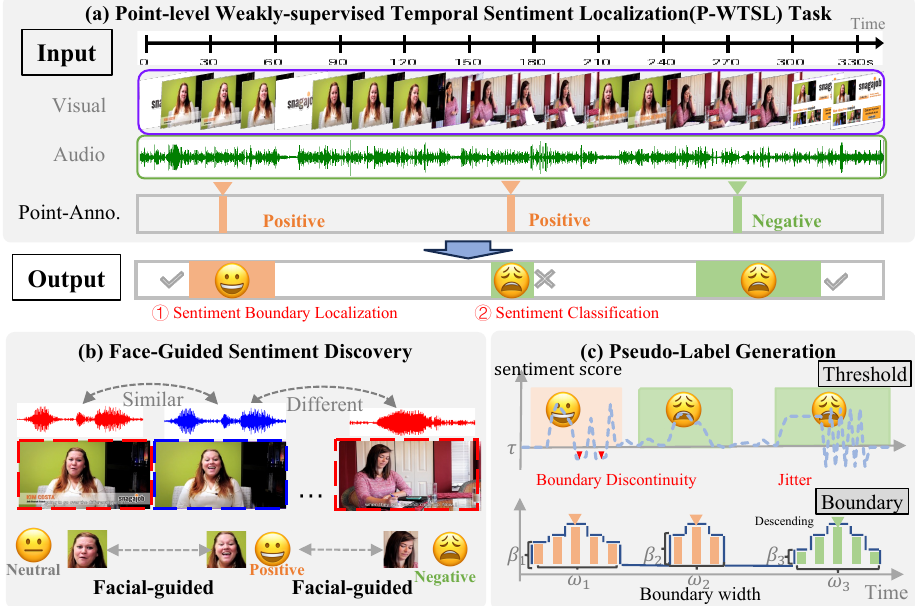}
\caption{\textbf{Illustration of the task and motivation.} (a) Given the point-level sentiment points in the untrimmed video, the model needs to achieve sentiment classification and boundary localization.
    (b) We explored the facial-centric method to guide sentiment clues discovery.
    (c) To alleviate sentiment boundary jitter and discontinuity under weak supervision, we propose a novel pseudo-label smoothing strategy.
}
\label{fig:enter-label}
\vspace{-1em}
\end{figure}

Driven by recent advances in audio-visual learning \cite{zhou2025open,zhao2025multimodal,li2024object,li2025patch,liu2025energy,zhou2025avs_semantics,zhou2026clasp,zhou2021positive,zhou2023contrastive,zhou2024label,zhou2024segment,zhou2022avs,jin2025simtoken,zhou2025think,guo2025audio,kurpath2025benchmark,shen2023fine,mao2024tavgbench,zhou2026audit,zhou2026mtavg}, Temporal Sentiment Localization (TSL), which aims to analyze sentiment in untrimmed audible videos, has become increasingly important for applications such as sentiment-adaptive editing~\cite{ji2021audio,wang2023reprompt}
and affect-aware summarization~\cite{apostolidis2021video,bertrand2022cognitive}.
However, acquiring dense frame-level sentiment annotations is time-consuming and costly, restricting the development of fully-supervised TSL~\cite{Poria2017Context, Zadeh2018MOSEI, majumder2018multimodal,yu2021learning, Tsai2019MultimodalTransformer}.
As a result, the weakly-supervised temporal sentiment localization (WTSL) is receiving more and more attention from the community.
Recent advances in WTSL mainly fall into two categories:
(1) \textbf{\textit{video-level weakly-supervision}}, where only entire video sentiment labels are available.
DCIN~\cite{li2022dilated} uses aligned subtitles as weak supervision via cross-modal consensus learning. Other weakly supervised methods include multiple instance learning ~\cite{wang2017untrimmednets,nguyen2018weakly}, reweighting-based mechanisms~\cite{hong2021crossmodal,lee2021weakly,zhou2025mettle,zhou2025dense}, and proposal scoring approaches~\cite{ren2023proposal,hu2023learning}. However, lacking fine-grained temporal labels, these methods often fail to precisely localize temporal boundaries.
(2) \textbf{\textit{point-level weakly-supervision}}, which uses a single point annotation per segment with the sentiment boundaries unknown.
Recent models TSL~\cite{zhang2022temporal} exploit temporal context and semantic alignment to infer sentiment segment localization.

In this research, we focus on point-level weakly-supervised TSL for videos, where the model aims to automatically locate fine-grained sentiment segments and classify their sentiment categories.
As shown in Fig.~\ref{fig:enter-label}(a), it constructs a mapping from sparsely annotated sentiment points and audio-visual modalities to segment-level sentiment boundary localization.
However, this task has some core challenges:
\textbf{1) Redundant Information Masks Salient Sentiment Signals}.
Videos contain diverse visual and audio content, but key sentiment-related information is often obscured by redundant content (\textit{e.g., `background' or `color'}), making it difficult for the model to mine effective sentiment information.
\textbf{2) Uncertain Sentiment Boundaries Due to Sparse Point Annotations}.
Point-level annotations in P-WTSL provide only a few sentiment anchors, making it difficult to determine precise sentiment boundaries.
\textbf{3) Abrupt and Transient Sentiment Changes}.
In realistic scenarios, sentiment may emerge and vanish abruptly within short temporal spans, which leads to significant challenges for accurate classification.

To address the above issues, we \textit{first} propose to enhance video temporal sentiment localization using the \textbf{Facial Features}, for the three reasons: \ding{172} Facial expressions directly reflect sentiment stimuli~\cite{sama2025attention, alp2024effect} in real scenarios; \ding{173} Subtle changes in the face area are easier to capture with lower difficulty, as shown in Fig.~\ref{fig:enter-label}(b), whereas the overall temporal differences in the main visual content, such as `\textit{characters}' and `\textit{background}', are more difficult to perceive due to redundancy. \ding{174} Face features are directly derived from visual frames and can better synergize learning with video features.
To this end, to effectively leverage facial features for temporal sentiment localization, we propose the \underline{Face-Guided Sentiment Boundary Enhancement Network} \underline{(FSENet)}.
Specifically, a face-centric interaction module is designed, consisting of two stages: the facial features first interact independently with audio and visual.
These relatively clean and unimodal features make it easier to detect sentiment cues.
To further capture more complex sentiment, the fused representations Face-Audio ($F_a^{(f)}$) and Face-Visual ($F_v^{(f)}$) are employed to interact for sentiment expression.
In parallel with the Facial-Centric Interaction branch, we incorporate a global sentiment perception weighting mechanism, which identifies potential sentiment-relevant segments from a holistic view. These two branches collaboratively leverage facial features to discover sentiment-stimulation cues for localization.

\begin{figure*}[!t]
    \centering
    \includegraphics[width=0.99\textwidth]{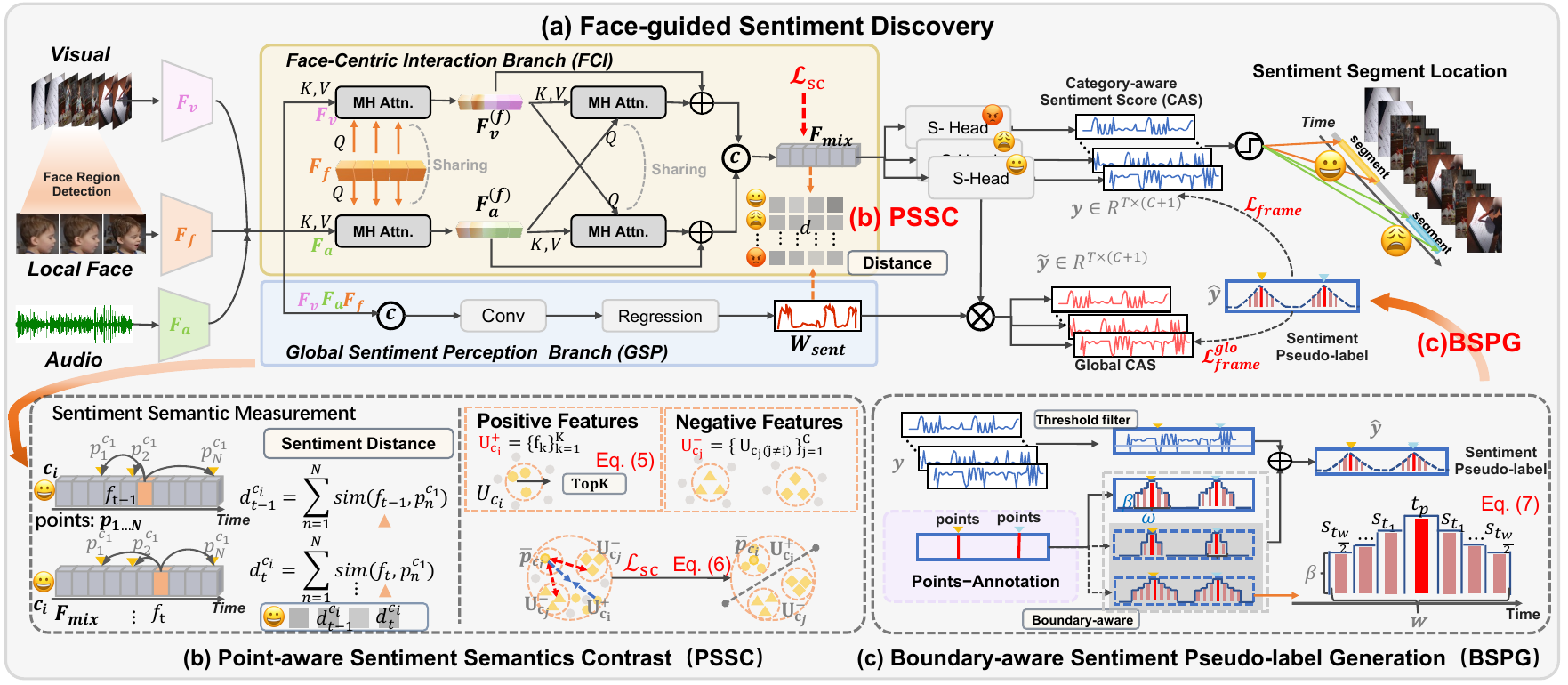}
    \captionof{figure}{Overview of our FSENet framework. (a)FSD module introduces \textbf{\textit{facial features}} into the temporal sentiment localization task through two branches to advance discovery of sentiment clues: \textit{one} focuses on facial-centric interaction, while \textit{the other} employs global sentiment perception weighting score. (b)PSSC aims to identify similar and dissimilar semantic \textit{points} in the temporal axis to enhance the \textit{\textbf{point-level sentiment discrimination}} by contrastive learning. (c)BSPG transforms \textit{sentiment point-level annotations} into \textit{\textbf{segment-level sentiment boundaries pseudo-label}} using the threshold filtering and step-by-step boundary-aware scheme, where sentiment scores decay progressively based on parameters $\beta$ and $w$.}
    \label{fig:pipeline}
\end{figure*}

Importantly, the point-level supervised TSL task still faces two key challenges: vague sentiment boundaries and poor segment-level classification after the stimulus cues are found. Therefore, we further propose \textbf{two key optimization modules} to enhance sentiment semantic alignment. The \underline{Point-aware Sentiment Semantics Contrast (PSSC)} (shown in Fig.~\ref{fig:pipeline}(b)) performs frame-level sentiment semantics contrastive training on the temporal axis.
Specifically, for each sentiment category (\eg, positive), we aggregate annotated positive point features into a class-specific prototype.
The top-$k$ frames most similar to the class-specific prototype are selected as positives ($\mathcal{U}^{+}$), while other unannotated frames serve as negatives ($\mathcal{U}^{-}$).
This contrastive training pulls positive frames closer in semantic space to improve segment sentiment classification.
Moreover, the proposed \underline{Boundary-aware Sentiment Pseudo-label Generation} \underline{(BSPG)} module aims to produce high-quality pseudo-labels for reliable sentiment boundary modeling under point-level supervision.
As illustrated in Fig.~\ref{fig:enter-label}(c), traditional threshold-based methods~\cite{lee2021learning} on frame-wise sentiment scores often suffer from {\textit{boundary discontinuity}} and {\textit{jitter}} due to score sensitivity and frame-level content noise, even after normalization.
To address this, BSPG introduces a label smoothing mechanism that gradually decays sentiment scores from each annotation point using a decay factor $\beta$, with boundary width $w$ controlling the extent of adjacent frames. This yields smoother, more coherent pseudo-labels, which are beneficial for determining effective sentiment boundaries.
Our main contributions are as follows.:
\begin{itemize}
    \item \textbf{Facial-Guided Multimodal Interaction for Sentiment Discovery.} We leverage facial features as key sentiment cues to guide audio-visual interaction, enabling interaction and global branches to discover more accurate sentiment cues amid complex background content.
    \item \textbf{Sentiment boundary optimization under weakly-supervised.} We design a contrastive scheme that pulls sentiment-consistent frames toward class prototypes derived from point annotations to enhance sentiment semantics consistency, and generate smoothing pseudo-labels by expanding annotation points into the temporal axis to improve boundary continuity and reduce noise.
    \item \textbf{Unified framework with strong generalization.} Our method provides a unified solution achieving state-of-the-art performance under full supervision and video-/point-level weak supervision, demonstrating strong generalization across varying settings, even with LLM methods.
\end{itemize}

\section{Related Work}
\subsection{Multimodal Sentiment Analysis}
Multimodal Sentiment Analysis (MSA) leverages visual, audio, and textual information to predict emotion for cropped videos~\cite{Zadeh2018MOSEI}.
Some approaches~\cite{hazarika2020misa,he2025domain,zhang2024training,song2024emotional} decompose features into shared and modality-specific subspaces to help capture consistent emotions across multimodal information.
Some work focus on the modality imbalance, filtering out noise from single modalities before fusion \cite{zeng2024multimodal, zhang2023learning} and removing noise from irrelevant frames to achieve better emotion recognition~\cite{zhang2023weakly}.
In addition, some work attempt to re-modify the prompt text to support the classification of emotion category~\cite{wei2024learning, chay2025llavac}.
However, current MSA methods are still limited to coarse classification of pre-cropped video clips and are incapable of handling untrimmed videos~\cite{singh2024survey}.
Our framework advances toward fine-grained multi-emotion understanding and temporal boundary localization in untrimmed videos.
\subsection{Temporal Boundary Localization}
Temporal boundary localization aims to detect the occurrence time of events in untrimmed videos. Common methods \cite{10632787,tang2023ddg, yun2024weakly, zhang2024hr} first generate a Class Activation Sequence and then apply a fixed threshold for boundary detection. However, these baseline approaches are primarily designed to learn better action representations for locating action boundaries, making them incapable of handling the subtle and dynamic nature of sentiment cues (e.g., facial expressions).
Recent Temporal Sentiment Localization studies~\cite{zhang2022temporal} use point supervision to jointly classify and localize sentiment segments, typically via inverse mapping and contrastive learning. However, they overlook temporal cross-modal dependencies(e.g., synchrony between facial expressions, visuals, and audio) and sentiment semantic opposition (e.g., positive vs. negative sentiment). In contrast, our method leverages facial emotion guidance to enhance cross-modal consistency and contrast between sentiment semantics, thereby significantly improving localization accuracy.

\section{Methodology}
\subsection{Problem Formulation}
Given an sentiment audible video with \( T \) frames, the point-level annotations along timeline are provided as \( Y_{\text{anno}} = \{ (t_i, y_{t_i}) \}_{i=1}^{N}\), where $i$-th point is denoted as $p_i$, and $p_i \in \mathbb{R}^C$ is a one-hot vector which notes the sentiment categories.
The point-level weakly-supervised temporal sentiment localization (P-WTSL) task aims to predict segments conveying sentiment around annotated points.

\subsection{Face-guided Sentiment Discovery (FSD)}
To effectively incorporate facial cues into temporal sentiment localization, we propose the Face-guided Sentiment Discovery(FSD) main module, which integrates facial cues to guide and enhance interaction with audio/visual modalities for sentiment temporal alignment.
FSD is designed with two parallel branches, Face-Centric Interaction (FCI) and Global Sentiment Perception (GSP) branches.
They cooperate with each other to assist in the discovery of sentiment.

\noindent\textbf{Face-Centric Interaction (FCI)} leverages facial embeddings as semantic guidance to selectively attend to emotionally salient segments, shown in Fig.~\ref{fig:pipeline}(a).
Prior approaches~\cite{li2022dilated,zhang2022temporal} generally overlook the role of facial features in TSL, but our strategy leverages facial embeddings to guide audio/visual feature interaction and highlight sentiment-relevant temporal segments.
Specifically, given facial feature $F_f$, visual feature $F_v$, and auditory feature $F_a$, where $F_f,F_a,F_v \in \mathbb{R}^{T \times d}$, the first stage face-guided interaction is formulated as follows:
\begin{equation}
    \begin{aligned}
        {F}_v^{(f)} &= F_v+\text{MHAttn}(F_f, F_v, F_v), \\
        {F}_a^{(f)} &= F_a+\text{MHAttn}(F_f, F_a, F_a), \label{eq:mh_face2audio} 
    \end{aligned}
\end{equation}
where the MHAttn($\cdot$) denotes the multi-head attention ~\cite{vaswani2017attention}, with its inputs corresponding to the \textit{`query'}, `\textit{key}', and `\textit{value}', respectively. The outputs ${F}_v^{(f)}$, ${F}_a^{(f)} \in \mathbb{R}^{T \times d}$ are face-guided fusion features in stage-I.

To further enable comprehensive sentiment clues discovery, we design the stage-II face-guided interaction to progressively integrate audio and visual features with the impact of facial features, which performs deep multimodal interaction.
The process is formulated as follows:
\begin{equation}
    \begin{aligned}
        {F}_v^{(af)} &=  {F}_v^{(f)} + \text{MHAttn}({F}_a^{(f)},F_v^{(f)}, F_v^{(f)}), \\
        {F}_a^{(vf)} &=  {F}_a^{(f)} + \text{MHAttn}({F}_v^{(f)},{F}_a^{(f)},{F}_a^{(f)}),
        \label{eq:mh_audio2face}
    \end{aligned}
\end{equation}
where the fusion output ${F}_v^{(af)}$ and ${F}_a^{(vf)} \in \mathbb{R}^{T \times d}$.
We concatenate them to obtain the fused representation $F_{\text{mix}}= [{F}_v^{(af)}; {F}_a^{(vf)}] \in \mathbb{R}^{T \times 2d}$.




\noindent\textbf{Global Sentiment Perception (GSP)} weighting strategy is designed to complement the interaction branch by modeling sentiment intensity.
While the Face-Centric Interaction branch enhances temporal alignment across modalities through facial feature guided interactions, this global branch takes a holistic view, identifying sentimentally salient segments along frame-level emotional stimulation dependencies.
Specifically, modality-specific features $F_a$, $F_v$, and $F_f$ are first concatenated, then fed into the convolutional layers $E(\cdot)$.
Then the output is projected by regression head $Reg(\cdot)$ with a sigmoid activation $\sigma(\cdot)$ to produce the sentiment weight \({W_{\text{sent}}} \in \mathbb{R}^{T \times {1}} \), and weighting \(W_{\text{sent}} \in [0,1]\):
\begin{equation}
    \begin{aligned}
     & W_{\text{sent}} = \sigma\left({Reg}\left(E\left([F_a; F_v; F_f]\right)\right)\right).
    \end{aligned}
\label{eq:sent-refine}
\end{equation}

\subsection{Point-aware Sentiment Semantics Contrast (PSSC)}
\label{sec:contrast}


To enhance the boundary localization of sentiment semantics around annotation points under weak supervision, we propose the PSSC optimization strategy, as illustrated Fig.~\ref{fig:pipeline}(b).
The capability of this strategy is to leverage point-level sentiment annotations to pull temporally points (at the frame-level) with the same sentiment closer in the embedding space.
Specifically, for each candidate point embedding $f_t$ on the temporal axis, we measure its distance to all annotated points $p_{1...n}^{c_i}$.
The computation is performed as follows:
\begin{equation}
    d_{t}^{c_i} = {\textstyle \sum_{n=1}^{N} \text{sim}(f_t, p_n^{c_i})},
\end{equation}
where \(\text{sim}(\cdot, \cdot)\) denotes the feature similarity (\ie, cosine similarity), and value $d \in [0,1]$, and the number $N$ means the annotated points of sentiment of $c_i$.

Next, we construct positive and negative sample sets for contrastive learning, based on the computed distances.
To gain sentiment stimulation, we incorporate Global Sentiment Perception weighting (GSP) in Eq.~\ref{eq:sent-refine}, which can adjust sample selection based on sentiment saliency.
The positive neighbor set for category $c_i$, denoted as $\mathcal{U}_{c_i}$, consists of the top-$K$ most relevant point embeddings:
\begin{equation}
\mathcal{U}^{+}_{c_i} = \{ f_t \mid d_t^{c_i} \in \text{Top-}K(\{W_{\text{sent},t} \cdot d_t^{c_i}\}_{t=1}^{T}) \},
\label{eq:topk}
\end{equation}
where $K<T-N$ is a parameter controlling the size of the positive set $\mathcal{U}^{+}_{c_i}$, which consists of point-level embeddings, and $t \neq t_p$ ($t_p$: annotated timestamp).
Meanwhile, for other sentiment categories $c_j \neq c_i$, nearby points form negative sets $\mathcal{U}^{-}_{c_j}$, as they belong to different sentiment prototypes.

To enhance the discriminative ability for sentiment classification under point-level supervision, we introduce a contrastive learning loss to encourage candidate points to be close to their annotation points with the same sentiment category, while simultaneously pushing them away from neighbors of different categories.
To this end, we construct a triplet $\{\bar{p}_{c_i}, \mathcal{U}^{+}_{c_i}, \mathcal{U}^{-}_{c_j}\}$ for sentiment $c_i$.
The $\bar{p}_{c_i}$ serves as a sentiment prototype, encouraging features in the positive set $\mathcal{U}^{+}_{c_i}$ to be pulled closer to $\bar{p}_{c_i}$, while pushing features in the negative sets $\mathcal{U}^{-}_{c_j}$ away.
The overall loss across all sentiment categories is defined as:
\begin{equation}
\mathcal{L}_{\text{sc}} = -\displaystyle \sum_{i=1}^{C} log{\frac{
  \textstyle \sum_{f_t \in \mathcal{U}^{+}_{c_i}} 
  exp(f_t \cdot \bar{p}_{c_i})
  }{
    \textstyle \sum_{f_t \in \{\mathcal{U}^{+}_{c_i}, \mathcal{U}^{-}_{c_j} \}} 
    exp(f_t \cdot \bar{p}_{c_{k=i/j}})
  }},
\label{eq:sc_loss}
\end{equation}
where $\bar{p}_{c_i}$ is the average result of the annotation point features for sentiment category $c_i$, and $k$ in negative sets is based on the sentiment to which $f_t$ belongs.




\subsection{Boundary-aware Sentiment Pseudo-label Generation (BSPG)}
\label{sec:step-pseudo}
The quality of the generated pseudo-label $\hat{\bm{y}}$, especially the accuracy of its sentiment boundaries, plays a pivotal role in the effectiveness of the point-level WTSL task.
To address the challenge of jitter and discontinuous pseudo-label distributions caused by using raw model predictions with a threshold filter~\cite{lee2021learning, rachavarapu2024weakly}.
We propose a \textit{Boundary-aware Sentiment Pseudo-label Generation} (BSPG) strategy that exploits the local temporal continuity of sentiment frames near annotation points.
The detailed procedure is as follows:
\begin{equation}
    \bm{s}_t = 
    \beta + (1-\beta)\left(1 - \frac{|t - t_p|}{w}\right),
\label{eq:beta_function}
\end{equation}
where $t_p$ is the timestamp of the annotated point, and $t$ is any time step along the video, and $w \in \mathbb{Z}^{+}$ and $\beta \in [0, 1]$ are the predefined hyperparameters denoting the smoothing window length (frames-level) and smoothing score that controls the minimum pseudo-label value at the temporal boundaries, respectively.

Next, we integrate the output sentiment score $\textbf{y}$ and the obtained smoothing scores $\bm{s}_t$, to enhance the reliability of sentiment localization supervision:
\begin{equation}
\resizebox{0.87\linewidth}{!}{$
    \hat{\bm{y}}_{t, c_i} = \mathbb{I}(\bm{y}_t)+
    \begin{cases}
    \bm{s}_t, & \text{if } c_i \neq c+1 \text{ and } |t - t_p| \le w, \\
    1 - \bm{s}_ t, & \text{if } c_i = c+1 \text{ and } |t - t_p| \le w, \\
    0, & \text{otherwise},
    \end{cases}
$}
\label{eq:width_function}
\end{equation}
where $c_i = c+1$ represents the non-sentiment category, and $\mathbb{I}(y_t)$ denotes a sentiment threshold function that sets the sentiment score to zero when non-sentiment confidence $\bm{y}_{t}$ exceeds threshold $\tau$=0.95, indicating a non-sentiment point.

\subsection{Optimization of Backbone Network}
To adapt the point-level weakly supervised temporal sentiment localization, we leverage the previously obtained $F_{\text{mix}}$ and $W_{\text{sent}}$ features to construct an effective loss function for optimizing our framework.
Specifically, the $F_{\text{mix}}$ is processed through $c+1$ independent convolutional layers with a sigmoid-activated function to generate the frame-level sentiment score for sentiment alignment supervision, and the process is as follows:
\begin{equation}
    \bm{y} = \sigma\left(\text{CLS}( F_{\text{mix}})\right) \in \mathbb{R}^{T \times {(C+1)}},
    \label{eq:cas}
\end{equation}
where the number $C$ denotes sentiment categories, with an additional category designated for non-sentiment.
The output $\bm{y}$ denotes the frame-level category-aware sentiment score (CAS).

Next, we further achieve feature-level collaboration between the two branches, incorporating the category-aware sentiment score (CAS) into the global sentiment perception (GSP) branch for joint computation:
\begin{equation}
    \tilde{\bm{y}}_t =  \left[ W_{\text{sent}, t} \cdot \bm{y}_{t, 1:C} \;; \;(1 - W_{\text{sent}, t}) \cdot \bm{y}_{t, C+1} \right],
\end{equation}
where the larger value in $W_{\text{sent}}$ reflect stronger sentiment stimuli in the first $c$ categories, meanwhile $1-W_{\text{sent}}$ is used to represent the last non-sentiment category, and \(\tilde{\bm{y}}_t \in \mathbb{R}^{C+1}\) is the sentiment-enhanced activation for frame \(t\), and \(\tilde{\bm{y}} \in \mathbb{R}^{T \times (C+1)}\) denotes the (Global CAS) in global branch.



Our optimization objective is to minimize the discrepancy between the category-aware sentiment score (CAS) and the global CAS with the sentiment pseudo label $\hat{\bm{y}}$ generated from point-level annotations.
To guide the training process, we define a base loss $\mathcal{L}_{\text{base}}=-\sum_{c=1}^{C} \text{avg}(\hat{\bm{y}}_{c}) \cdot \log \text{avg}(\bm{y}_{c})$ that encourages alignment of (video-level) sentiment predictions by averaging over temporal axis.
We also introduce two finer-grained (frame-level) supervision constraints, $\mathcal{L}_{\text{frame}}$ and $\mathcal{L}^{\text{glo}}_{\text{frame}}$, whose details are as follows:
\begin{equation}
\begin{aligned}
    &\mathcal{F}_{\text{align}}(\bm{f}; \bm{f}_p) =  \textstyle \sum_{c=1}^{C+1} \text{FL}(\bm{f}; \bm{f}_p), \\
&\underbrace{
\begin{aligned}
&\mathcal{L}_{\text{frame}} = \mathcal{F}_{\text{align}}(\bm{y}; \hat{\bm{y}});
&\mathcal{L}^{\text{glo}}_{\text{frame}} = \mathcal{F}_{\text{align}}(\tilde{\bm{y}}; \hat{\bm{y}}),
\end{aligned}
}_{\begin{array}{lll}
\bm{y} \leftarrow{} \text{FCI} & \tilde{\bm{y}} \leftarrow{}  \text{GSP} & \hat{\bm{y}}\leftarrow{} \text{BSPG }  
\end{array}}
\end{aligned}
\label{eq:frame_loss_joint}
\end{equation}
where the $\mathcal{F}_{align}$ ($\cdot$\ ; $\cdot$) function performs \textbf{F}rame-level \textbf{S}entiment \textbf{Align}ment between the model outputs ($\bm{y}$ and $\tilde{\bm{y}}$) and the pseudo labels ($\hat{\bm{y}}$).
The $\text{FL}$ denotes the focal loss function~\cite{lin2017focal}with $\gamma$ is $2$, which is widely used in multimodal sentiment analysis tasks to help distinguish sentiment-related frames from non-sentiment ones.


In summary, the overall backbone network is trained on the point-level WTSL task with multiple loss functions. The collaborative loss computation $\mathcal{L}_{\text{total}}$ is as follows:
\begin{equation}
\mathcal{L}_{\text{total}}=\mathcal{L}_{\text{base}} + \lambda_{1}\cdot(\mathcal{L}_{\text{frame}}+\mathcal{L}^{\text{glo}}_{\text{frame}}) + \lambda _{2}\mathcal{L}_{\text{sc}},
   \label{eq:loss_total}
\end{equation}
where the hyperparameters $\lambda_1$ and $\lambda_2$ are weighting factors that balance the relative importance of the respective terms in the total loss.

\section{Experiments}
\label{sec:experiments}

\begin{table*}[!t] 
  \centering
  \setlength\tabcolsep{8pt}
  \caption{\textbf{Performance comparison with SOTA methods under Fully and Weakly supervised settings on TSL300}. ``$\dagger$'' indicates methods reproduced by ourselves. The best and second-best results are \textbf{bolded} and \underline{underlined}, respectively.}
  \renewcommand\arraystretch{0.95}
  \resizebox{0.95\textwidth}{!}{
  \begin{tabular}{c|c|ccccc|ccc} 
    \toprule
    \multirow{3}{*}{\makecell{\textbf{Setting}}} & \multirow{3}{*}{\textbf{Method}} & \multicolumn{5}{c}{\textbf{mAP@IoU(\%)}} &\multicolumn{3}{c}{\textbf{Average}} \\ 
    \cmidrule(lr){3-7} \cmidrule(lr){8-10}
     &  & \textbf{0.1} & \textbf{0.15} & \textbf{0.2} & \textbf{0.25} & \textbf{0.3} &\textbf{Recall} & \textbf{F2 Score} & \textbf{mAP} \\ 
    \midrule
    \multirow{3}{*}{\makecell{\textbf{Full Supervision}}}
    & VAANet~\cite{zhao2020end} & 28.01 &24.95 & 20.68 & 16.45 & \underline{12.50} &69.82& \textbf{39.59} & 20.51 \\ 
     & $\dagger$AFSD~\cite{lin2021learning}& \textbf{33.44} &\textbf{27.09} & \underline{21.78} & \underline{16.64} & 11.76& \underline{75.10} & \underline{31.78} & \textbf{22.14} \\
     &\textbf{FSENet(ours)} &\underline{28.02}&\underline{25.45}&\textbf{22.63}&\textbf{18.93}&\textbf{14.33}&\textbf{79.58}&30.37&\underline{21.87}\\ 
    \midrule
    \multirow{6}{*}{\makecell{\textbf{Weak Supervision} \\ \textbf{(Video-level)}} }
     & CoLA~\cite{zhang2021cola} & 11.87 & 10.14 & 8.11 & 6.83 & 4.98 &\underline{64.07}& 21.19 & 8.38 \\ 
     & UM~\cite{lee2021weakly}& 14.77 & \underline{13.28}& \underline{10.77}& 8.41 & 6.54 &55.33 & 23.94 & \underline{10.75} \\ 
     & $\dagger$DDG-Net~\cite{tang2023ddg}& \underline{16.13}& 12.19& 9.91& \underline{8.47}& \underline{6.93}&\textbf{66.38}& 23.82& 10.72\\ 
     & $\dagger$ISSF~\cite{yun2024weakly}& 13.83& 11.15& 9.30& 6.87& 5.14&45.65& 25.17& 9.26\\ 
      & $\dagger$TSL~\cite{zhang2022temporal}& 14.10& 11.35& 8.77& 6.36& 4.47&31.84& \underline{27.40}& 9.01\\
    & \textbf{FSENet(ours)}& \textbf{19.15}& \textbf{15.12}& \textbf{11.91}&\textbf{9.35}&\textbf{7.12}&58.11&\textbf{30.0}&\textbf{12.53}\\ 
    \midrule
    \multirow{6}{*}{\makecell{\textbf{Weak Supervision} \\ \textbf{(Point-level)}}}
     & LACP~\cite{lee2021learning}& 17.25 & 14.46 & 12.47 & 9.59 & 7.76 & 66.92 & 25.01 & 12.30 \\ 
     & SF-Net~\cite{ma2020sf}& 20.26 & 18.09 & 15.01 & 12.38 & 9.76 &59.55 & 32.61 & 15.10 \\ 
      & $\dagger$\text{HR-pro}\textsuperscript{stage1}~\cite{zhang2024hr} & 19.77 & 17.10 & 14.71 & 11.38 & 8.86 &62.23 & 22.82 & 14.36 \\ 
     & $\dagger$\text{{HR-pro}\textsuperscript{stage2}}~\cite{zhang2024hr}
     & 27.24 & 23.13 & 19.85 & 15.96 & \underline{11.83} &\underline{73.60} & 32.21 & 19.60\\  
     & TSL~\cite{zhang2022temporal} &\underline{28.72} & \underline{24.92} &\underline{20.46} & \underline{16.10} & \underline{11.83} &71.14 &\textbf{35.36}& \underline{20.40} \\ 
     & \textbf{FSENet (ours)} & \textbf{29.31} & \textbf{25.47} & \textbf{22.49} & \textbf{16.76} & \textbf{13.24} &\textbf{75.02} & \underline{33.67} & \textbf{21.45} \\ 
    \bottomrule
  \end{tabular}
  }
\label{tab:method_comparison}
\end{table*}
\subsection{Experimental Setup}

\textbf{Dataset and Evaluation Metrics}.
We primarily evaluate our method on the popular point-annotated TSL300~\cite{zhang2022temporal} dataset, which contains 300 untrimmed videos with an average duration of over 250 seconds, totaling more than 1.3$k$ minutes. We also evaluate our method on CMU-MOSEI~\cite{zadeh2018multimodal}, where we retain two sentiment categories for temporal localization on untrimmed videos. Following the standard protocol~\cite{zhang2022temporal}, we use mean Average Precision (mAP) under IoU thresholds from 0.1 to 0.3 (step size 0.05) as the primary metric for temporal localization.
Additionally, we report Recall and F2-score metrics to better evaluate segment-level sentiment classification.

\noindent\textbf{Implementation Details.}
Videos are divided into 16-frame segments to extract visual, audio, and facial features using the I3D~\cite{carreira2017quo}, MFCC~\cite{muda2010voice}, and ResEmoteNet~\cite{roy2024resemotenet}, respectively. Facial regions are detected using DeepFace\footnote{\url{https://github.com/serengil/deepface}} to obtain bounding boxes: [x,y,w,h], which are then fed into ResEmoteNet for feature extraction.
The raw feature dimensions (1024/512/256) are projected to a unified dimension $d$=512 via unfixed 1D convolutions.
We use the Adam optimizer, with an initial learning rate of \( 1 \times 10^{-5} \) and a weight decay of \( 5 \times 10^{-4} \).
The training epochs are 600, and the batch size is 16.
The experiments are conducted on a single NVIDIA A40 GPU.
The optimization of total losses contains two parameters: $\lambda_1$=1 and $\lambda_2$=0.05.
The hyperparameters $w$=7 and $\beta$=0.6 are applied in Eq.~\ref{eq:beta_function} and Eq.~\ref{eq:width_function}, respectively.
\textit{\textbf{Supervision settings}}:
We compare point-level, video-level weak, and full supervision settings for TSL.
For weak supervision, only point annotations are used.
For the video-level setting, points are converted into a binary vector $\mathbb{R}^{C}$, where 1 indicates the presence of sentiment segments.
For full supervision, segment start and end times are converted into frame-level labels.
\begin{table}[!t]
\centering
\setlength\tabcolsep{6pt}
\caption{\textbf{Performance results on CMU-MOSEI dataset} under point-level settings.}
\resizebox{0.99\columnwidth}{!}{
\begin{tabular}{l|ccc|ccc}
\toprule
\multirow{2}{*}{\textbf{Method}} & \multicolumn{3}{c|}{\textbf{mAP@IoU(\%)}} & \multicolumn{3}{c}{\textbf{Average}} \\
\cline{2-7}
~ & \textbf{0.1} & \textbf{0.2} & \textbf{0.3} & \textbf{Recall} & \textbf{F2 Score} & \textbf{mAP} \\  
\midrule
{HR-pro}\textsuperscript{stage1} & 11.83 & 7.26 & 4.44 & 18.81&18.46&7.65 \\
{HR-pro}\textsuperscript{stage2} & 16.27  & 11.49  & 6.76 & 26.61&24.71&11.47 \\
TSL & 15.18 & 10.71 & 7.46 & 27.24 & 24.86 & 11.00 \\
\textbf{FSENet (Ours)} & \textbf{22.7} & \textbf{22.49} & \textbf{13.24} & \textbf{33.08} & \textbf{24.89} & \textbf{16.54} \\
\bottomrule
\end{tabular}}
\label{tab:cmu_mosei_results}
\end{table}

\begin{table}[t]
\centering
\setlength\tabcolsep{6pt}
\caption{\textbf{Comparison with large language models (LLMs)} on the TSL300 dataset. Zero-shot methods are evaluated without fine-tuning. while LoRA-based models are fine-tuned with point-level supervision.
}
\resizebox{0.99\columnwidth}{!}{
\begin{tabular}{l|ccc|c}
\toprule
\multirow{2}{*}{\textbf{Method}} & \multicolumn{3}{c|}{\textbf{mAP@IoU (\%)}} & \multirow{1}{*}{\textbf{Avg}} \\
\cline{2-4}
~ & 0.1 & 0.2 & 0.3 & \textbf{mAP}\\  
\midrule
Qwen3-Omni(zero-shot) & 7.67 & 5.03 & 2.67 & 5.07  \\
LLaMA-2-7B (LoRA) & 10.98 & 10.98 & 8.45 & 9.97  \\
LLaMA-2-7B (LoRA+BSPG) & 27.76 & 11.82 & 5.16 & 11.75 \\
\textbf{Ours (small model)} & 29.31 & 22.49 & 13.24 & 21.45 \\
\bottomrule
\end{tabular}}
\label{tab:temporal_emotion}
\end{table}

\begin{table}[!t]
  \centering
  \caption{\textbf{Fair evaluation under identical feature settings on TSL300}. Group \ding{192} uses only audio-visual features, and group \ding{193} adds local facial features. All $^{\triangle}$ adopt the same facial extractor.}
  \renewcommand\arraystretch{0.9}
  \resizebox{0.99\columnwidth}{!}{
  \begin{tabular}{l|ccccc|c}
    \toprule
    \multirow{2}{*}{\textbf{Method}}& \multicolumn{5}{c|}{\textbf{mAP@IoU (\%)}} & \textbf{Avg} \\
    & 0.1 & 0.15 & 0.2 & 0.25 & 0.3 & \textbf{mAP} \\ 
    \hline
    \rowcolor{gray!20}
    \multicolumn{7}{c}{\textit{AV Features Only}} \\
    \hline
    HR-pro\textsuperscript{stage1} & 16.64 & 13.95 & 11.86 & 9.74 & 8.01 & 12.04 \\
    HR-pro\textsuperscript{stage2} & 26.40 & 21.80 & 16.91 & 13.27 & 9.96 & 17.67 \\
    TSL& 28.14 & 23.02 & 18.68 & 14.36 & 10.46 & 18.93 \\
    FSENet(ours) & 25.91 & 24.70 & 20.08 & 16.85 & 11.53 & 19.41 \\
    \hline
    \rowcolor{gray!20}
    \multicolumn{7}{c}{\textit{AV + Facial Features}} \\
    \hline
    $^{\triangle}$HR-pro\textsuperscript{stage1} & 19.77 & 17.10 & 14.71 & 11.38 & 8.86 & 14.36 \\
    $^{\triangle}$HR-pro\textsuperscript{stage2} & 27.24 & 23.13 & 19.85 & 15.96 & 11.83& 19.60 \\
    $^{\triangle}$TSL                           & 27.77 & 24.94 & 20.93 & \textbf{17.15} & \textbf{13.48} & 20.86 \\
    $^{\triangle}$ FSENet(ours)                                & \textbf{29.31} & \textbf{25.47} & \textbf{22.49} & 16.76 & 13.24 & \textbf{21.45} \\
    \bottomrule
  \end{tabular}
  }
\label{tab:prior_works}
\end{table}

\subsection{Comparison to Prior Work}
We evaluate our model under multiple supervision settings, including \underline{full}, \underline{video-level}, and \underline{point-level} supervision, to assess its effectiveness on the temporal sentiment localization task across the TSL300 and CMU-MOSEI datasets.

\noindent \textbf{TSL300}: As shown in Tab.~\ref{tab:method_comparison}, under point-level setting, our model achieves superior performance, attaining the best mAP scores across all IoU thresholds [0.1:0.3], and the highest average mAP@IoU of 21.45\%, outperforming the previous SOTA method TSL~\cite{zhang2022temporal} by over 5\%, and achieving a peak Recall of 75.02\%. Moreover, our model (FSENet) also demonstrates competitive results under both full and video-level weak supervision. First, our method surpasses the video-level supervised SOTA methods, such as UM~\cite{lee2021weakly}, by 25\% in F2-score and 16.5\% in average mAP. Next, FSENet achieves a mAP of 21.87\% and an F2-score of 30.37\% under full supervision, approaching the performance of AFSD~\cite{lin2021learning}, which is specifically designed for the fully supervised setting. Our FSENet still surpasses this fully supervised method by around 6\% in Recall.

\noindent
\textbf{CMU-MOSEI:}
To further verify generalization, we evaluate FSENet on the CMU-MOSEI dataset. In Tab.~\ref{tab:cmu_mosei_results}, FSENet achieves mAP@IoU thresholds of 22.7\%, 22.49\%, 13.24\% for 0.1, 0.2, and 0.3, respectively. Despite the overall lower performance due to coarse temporal annotations, FSENet achieves 16.54\% average mAP, outperforming TSL and HR-pro by 5.54\% and 5.07\%, respectively.

\noindent
\textbf{Comparison with LLMs}. We compare our model with large language models (LLMs). Despite their strong reasoning ability, LLMs' text-centric representations hinder temporal sentiment alignment across audio/visual/facial modalities. As shown in Tab.~\ref{tab:temporal_emotion}, our model achieves the highest average mAP of 21.45\%, outperforming Qwen3-Omni-30B~\cite{xu2025qwen3omni}, LLaMA-2-7B~\cite{touvron2023llama2} with LoRA~\cite{hu2022lora}, and LoRA + BSPG by 16.30\%, 11.48\%, and 9.70\%, respectively.
Moreover, for LLaMA-2 fine-tuning, our pseudo-labeling strategy BSPG (Eq.~\ref{eq:width_function}) improves average mAP by 1.8\% over the variant without BSPG, demonstrating its effectiveness for temporal sentiment detection.
\begin{table}[!t]
\centering
\renewcommand\arraystretch{0.9}
\caption{{Ablation study on the main network component in FSD}. We report the result of mAPs, and `--' denotes not used.}
\resizebox{0.99\columnwidth}{!}{
\begin{tabular}{c|c|c|c|ccc|c}
\toprule
\multirow{2}{*}{\#id} & \multicolumn{2}{c|}{\textbf{FCI}} & \multirow{2}{*}{\textbf{GSP}} & \multicolumn{3}{c|}{\textbf{mAP@IoU (\%)}} & \multirow{1}{*}{\makecell{\textbf{Avg}}}  \\
\cline{2-3} \cline{5-7}
~ & first stage & second stage & ~ & 0.1 & 0.2 & 0.3 & \textbf{mAP} \\
\midrule
a &--   &--    & \checkmark & 24.95 & 18.63 & 11.52 & 18.42 \\ 
b & $\checkmark$ & --    & \checkmark & 27.19 & 18.93 & 11.16 & 19.10 \\ 
c & --   & $\checkmark$ & \checkmark & 27.46& 18.31 & 10.37 & 18.78 \\  
d & $\checkmark$ & $\checkmark$ & --    & 28.69 & 20.83 & 11.26 & 20.22\\  
e & $\checkmark$ & $\checkmark$ & $\checkmark$ & \textbf{29.31} &\textbf{22.49} & \textbf{13.24} & \textbf{21.45} \\  
\bottomrule
\end{tabular}}
\label{tab:fusion_strategy_detail}
\end{table}

\begin{table}[!t]
\centering
\setlength\tabcolsep{4pt}
\caption{Ablation of pseudo-label generation strategies.}
\renewcommand\arraystretch{0.9}
\resizebox{0.99\columnwidth}{!}{
\begin{tabular}{l|c|c|ccc|c}
\toprule
\multirow{2}{*}{\textbf{Pseudo-Label}} & \multirow{2}{*}{Threshold} & \multirow{2}{*}{Boundary} & \multicolumn{3}{c|}{\textbf{mAP@IoU(\%)}} & \textbf{Avg} \\
\cline{4-6}  
~ & ~ & ~ & 0.1 & 0.2 & 0.3 & \textbf{mAP} \\  
\hline
None                       & -- & --       &27.51&15.31&8.72  & 16.92  \\
Threshold Only            & \checkmark & --       &28.06&20.98&11.55 & 20.29\\
Step Only                 & --         &\checkmark    &25.76&21.59&13.10  & 20.31 \\
\textbf{Ours (full model)}      & \checkmark &\checkmark   & \textbf{29.31}&\textbf{22.49}&\textbf{13.24}  & \textbf{21.45} \\
\bottomrule
\end{tabular}}
\label{tab:pseudo-label}
\end{table}

\subsection{Ablation Studies}
\label{sec:ablation}

\noindent\textbf{Ablation studies on facial features}. To evaluate the impact of facial features, we conduct two group experiments: 1) using only audio-visual (AV) features, and 2) enhancing AV features with local facial features.
As shown in Tab.~\ref{tab:prior_works}, our method outperforms the SOTA method by 0.48\% even without facial features.
With facial feature enhancement, the performance of FSENet improves consistently across all IoU thresholds [0.1:0.3], reaching an average mAP of 21.45\% and outperforming $^{\triangle}$TSL~\cite{zhang2022temporal} by 0.59\%. These results indicate that facial features improve sentiment localization, and FSENet achieves the best performance across all settings, confirming the generalizability of our design.

\noindent \textbf{Effects of Main Network Components.}
The FSD framework consists of two parallel components: FCI and GSP.
We conduct an ablation study to evaluate their contributions (Tab.~\ref{tab:fusion_strategy_detail}).
Compared to (row \#a), (rows \#b and \#c) show slight improvements of 0.7\% and 0.4\%, respectively, when facial feature guidance is introduced into FCI.
However, directly applying the complex second-stage fusion of $F^{(f)}_a$ and $F^{(f)}_v$ (Eq.~\ref{eq:mh_audio2face} in (row \#c) does not lead to obvious improvements.
The best performance is achieved with an Avg.mAP of 21.45\% (row \#e), where facial guidance is progressively introduced (including both stages).
Moreover, when the GSP branch is ablated (from row \#e to \#d), the performance drops by around 1.2\%, highlighting the effectiveness of both the proposed FCI and GSP modules in leveraging facial features to enhance sentiment localization.

\noindent \textbf{Effects of the generated pseudo-label}.
Our Boundary-aware Sentiment Pseudo-label Generation (BSPG) strategy combines a traditional threshold-based approach with a novel point-centric, step-wise boundary generation method (Tab.~\ref{tab:pseudo-label}).
The naive baseline using raw predictions suffers from low-confidence noise (16.92\% mAP).
Applying a confidence threshold improves the result to 20.29\%, while our boundary-aware pseudo-label method refinement temporal consistency and further boosts mAP to 20.31\%.
Our full approach (BSPG) achieves the best result (21.45\% mAP) and improvement ($\approx$ 4.5\%), demonstrating that our joint design effectively facilitates sentiment boundary localization under weak supervision.

\begin{table}[!t]
\centering
\caption{{Ablation of weakly-supervised optimization losses}.}
\setlength\tabcolsep{10pt}
\renewcommand\arraystretch{0.9}
\resizebox{0.99\columnwidth}{!}{
\begin{tabular}{c|ccc|ccc|c}
\toprule
\multirow{2}{*}{\#id}  & \multirow{2}{*}{$\mathcal{L}_{\text{frame}}$} & \multirow{2}{*}{$\mathcal{L}_{\text{frame}}^{glo}$}  & \multirow{2}{*}{$\mathcal{L}_{\text{sc}}$} 
& \multicolumn{3}{c|}{\textbf{mAP@IoU(\%)}} & \multirow{1}{*}{\textbf{Avg}} \\
\cline{5-7}
 & & & & 0.1 & 0.2 & 0.3 &\textbf{mAP} \\
\midrule
a & \checkmark & \checkmark &\checkmark & \textbf{29.31}& \textbf{22.49}& \textbf{13.24}& \textbf{21.45}\\
\midrule
b &--  &\checkmark & \checkmark& 28.28 & 19.52 & 10.33& 20.02 \\
c & \checkmark &--  & \checkmark& 27.67 & 20.40 & 9.99& 20.65 \\
d & \checkmark & \checkmark &-- & 29.17 & 21.18 & 12.49 & 20.94 \\
\midrule
e &-- &-- &\checkmark& 19.15& 11.91&7.12&12.53 \\
f &-- &-- &-- & 16.61 & 12.32 &  9.38 & 10.69 \\
\bottomrule
\end{tabular}}
\label{tab:loss_ablation}
\end{table}

\begin{figure*}[!t]
    \centering
    \includegraphics[width=0.99\textwidth]{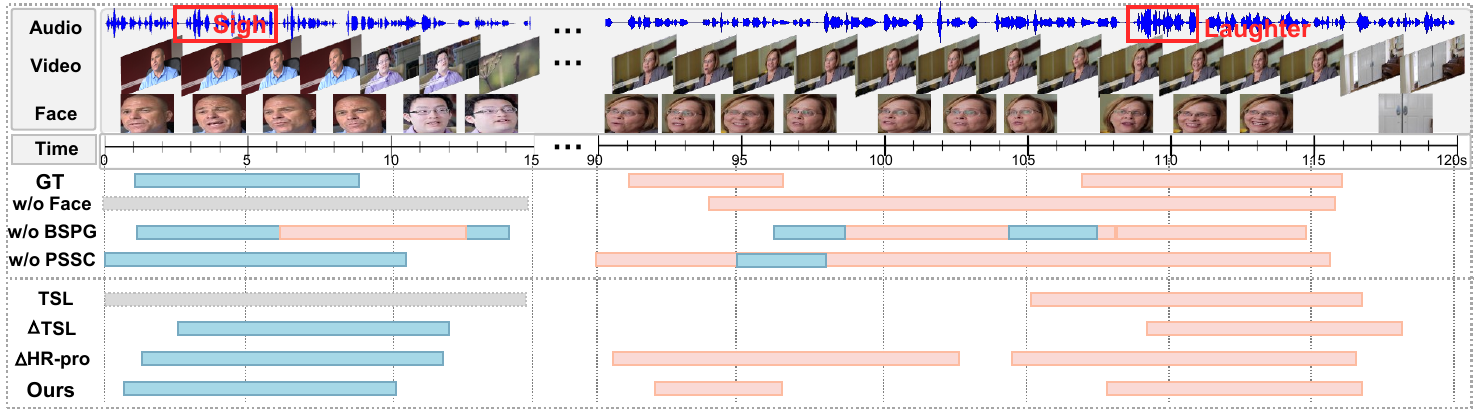}
    \caption{Visualization of the predicted results for temporal sentiment localization. The `pink' segments are positive sentiment, `blue' means negative, and `gray' means no valid location output. $\triangle$ indicates that methods incorporate facial features settings.}
    \label{fig:visual}
\end{figure*}

\begin{figure}[!t]
    \centering
    \includegraphics[width=0.99\columnwidth]{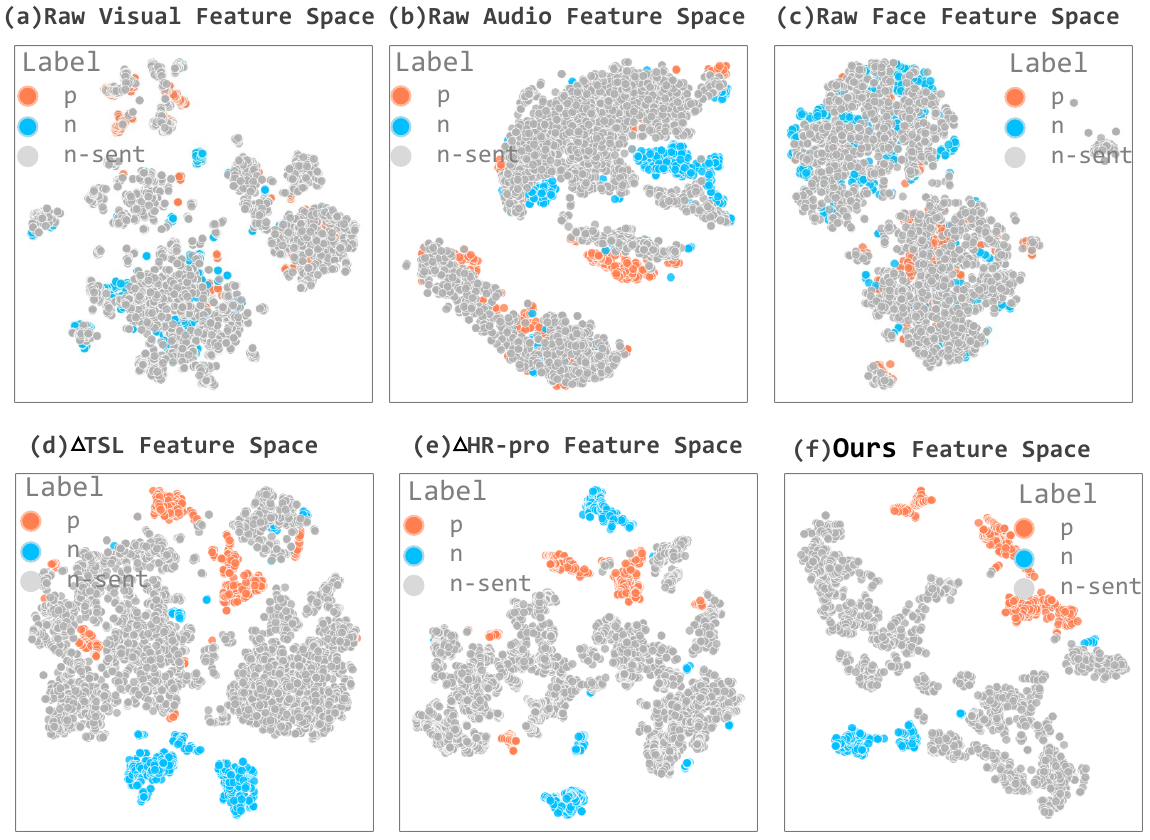}
    \caption{T-SNE visualization comparing our FSENet with SOTA. Each point represents a segment-level feature (‘p/n/n-sent’ denotes positive, negative, and non-sentiment categories, respectively).}
    \label{fig:feature_vis}
\end{figure}

\noindent \textbf{Effects of the optimization.}
The losses $\mathcal{L}_{\text{frame}}$ and $\mathcal{L}^{\text{glo}}_{\text{frame}}$ (Eq.~\ref{eq:frame_loss_joint}) enforce sentiment boundary constraints between the outputs of the two branches and the pseudo-labels under weak supervision. The semantic contrast loss $\mathcal{L}_{sc}$ (Eq.~\ref{eq:sc_loss}) further aligns annotation points with nearby sentiment-relevant frames.
We conduct an ablation study on these optimization losses, as shown in Tab.~\ref{tab:loss_ablation}. Compared with the full optimization setting (row \#a), removing each optimization loss (\#b/\#c/\#d) leads to performance drops of around 1.4\%, 0.8\%, and 0.5\%, respectively.
Notably, jointly removing $\mathcal{L}_{\text{frame}}$ and $\mathcal{L}^{\text{glo}}_{\text{frame}}$ causes a significant drop of 8.9\%, demonstrating the effectiveness of our pseudo-label design for sentiment boundary constraints. Overall, the results confirm that all proposed optimization losses contribute to improving temporal sentiment localization.

\subsection{Qualitative Results}
\textbf{Visualization of predicted sentiment localization}.
As illustrated in Fig.~\ref{fig:visual}, we present a long-form interview discussing U.S. student loans, involving three speakers whose sentiments shift from initially negative to increasingly positive. 
The results are grouped for analysis. The \textit{first} four outputs correspond to ablations of our main modules.
Without facial features (w/o Face), the model fails to detect the early negative sentiment segment (0--15s) due to the lack of facial guidance.
Removing our proposed optimization strategies (w/o BSPG and w/o PSSC) leads to fragmented sentiment predictions between (90--116s), indicating weaker temporal coherence and boundary precision.
The latter four outputs compare our method with existing methods.
TSL~\cite{zhang2022temporal} and $\triangle$TSL miss the fine-grained positive segment from (90--96s), suggesting limited sensitivity to short-duration sentiment shifts.
Although $\triangle$HR-pro~\cite{zhang2024hr} detects all emotional segments, it tends to overextend short segments into overly long ones, leading to boundary overestimation.
Our FSENet achieves the most accurate and balanced results, effectively capturing both short and long sentiment segments by leveraging facial cues and robust boundary strategies.

\noindent \textbf{T-SNE visualization of the sentiment points distribution.}
To evaluate the effectiveness of our model in learning sentiment semantics, we randomly select four videos from TSL300~\cite{zhang2022temporal} and apply t-SNE to their frame-level features.
Fig.~\ref{fig:feature_vis}(a)--(c) show the feature spaces of the original visual, audio, and facial features, where points with the same sentiment label are not clearly clustered. We then compare several state-of-the-art methods, including $\triangle$TSL~\cite{zhang2022temporal} and $\triangle$HR-pro~\cite{zhang2024hr}, as shown in Fig.~\ref{fig:feature_vis}(d)--(e).
The t-SNE results show multiple sentiment clusters for both positive and negative samples, indicating a limited ability to distinguish sentiments. In contrast, our method, FSENet (Fig.~\ref{fig:feature_vis}(f)), produces two well-separated clusters corresponding to positive and negative sentiments with a large inter-cluster distance, facilitating more accurate temporal sentiment localization.
This demonstrates the strong capability of our method in discriminating sentiment semantics.

\section{Conclusion}

In this paper, we propose FSENet, a unified framework for point-level weakly-supervised temporal sentiment localization (P-WTSL).
We attempt to enhance sentiment stimuli discovery by introducing facial features as core semantic guidance (FSD), which facilitates interaction between audio/visual and facial features.
Through facial-guided interaction, our model enhances sentiment boundary perception by jointly performing semantic contrastive learning (PSSC) and generating boundary-aware pseudo-labels (BSPG) that provide sentiment temporal supervision.
Comprehensive experiments and ablation analyses confirm that FSENet achieves superior performance to existing SOTA approaches under every supervision setting for TSL tasks.

\noindent
\textbf{Acknowledgement}:This work is supported by the National Key R\&D Program of China (NO.2024YFB3311600), the Natural Science Foundation of China (62272144), the Anhui Provincial Natural Science Foundation (2408085J040), the Major Project of the Anhui Provincial Science and Technology Breakthrough Program (202423k09020001), and the Fundamental Research Funds for the Central Universities (JZ2024AHST0337).
{
    \small
    \bibliographystyle{IEEEtran}
    \bibliography{main}
}

\clearpage
\setcounter{page}{1}
\maketitlesupplementary
\setcounter{section}{0}
\renewcommand{\thesection}{\arabic{section}}
\renewcommand{\thesubsection}{\thesection.\arabic{subsection}}
\renewcommand{\thesubsubsection}{\thesubsection.\arabic{subsubsection}}

\section{Hyperparameter Ablation}
\noindent\textbf{Ablation on BSPG Parameters.} 
Our BSPG improves the reliability of pseudo-labels under point-level supervision through step-by-step smoothing. This reduces temporal jitter and mitigates boundary discontinuities (Fig.~\ref{fig:pipeline}(c), Eq.~\ref{eq:beta_function}). 
We ablate two key hyperparameters: the smoothing width $w$ and the confidence decay factor $\beta$. 
As shown in Table~\ref{tab:multi_parameter_analysis}, performance peaks at $w = 7$ (mAP: 21.45\%), indicating that moderate neighborhood expansion effectively aggregates supervisory cues. Smaller windows (w = 5, mAP: 19.22\%) under-smooth labels, whereas larger ones (w = 9, mAP: 19.26\%) over-smooth and blur temporal boundaries. 
For the decay factor, optimal performance is achieved at $\beta = 0.6$. Lower values ($\beta$ = 0.1, mAP: 19.64\%) decay too quickly, limiting contextual information, whereas higher values ($\beta$ = 0.7, mAP: 21.20\%) overconfidently propagate pseudo-labels, introducing spurious activations. 
Overall, tuning both $w$ and $\beta$ improves pseudo-label quality and temporal localization.

\begin{table}[t]
\centering
\setlength{\tabcolsep}{4pt} 
\caption{Ablation study of BSPG parameters (confidence decay factor \(\beta\) and smoothing width \(w\)) on mAP performance.}
\label{tab:multi_parameter_analysis}
\begin{tabular}{l|cccccc}
\toprule
\multirow{4}{*}{\centering BSPG} & $w$ & 5 & 6 & 7 & 8 & 9 \\
&mAP (\%) & 19.22 & 19.77 & \textbf{21.45} & 19.66 & 19.26 \\
\cmidrule{2-7}
 &$\beta$ & 0.1 & 0.3 & 0.5 & 0.6 & 0.7 \\
&mAP (\%) & 19.64 & 18.88 & 21.29 & \textbf{21.45} & 21.20 \\
\bottomrule
\end{tabular}
\end{table}

\begin{figure}[htp]
    \centering
    \includegraphics[width=0.8\linewidth]{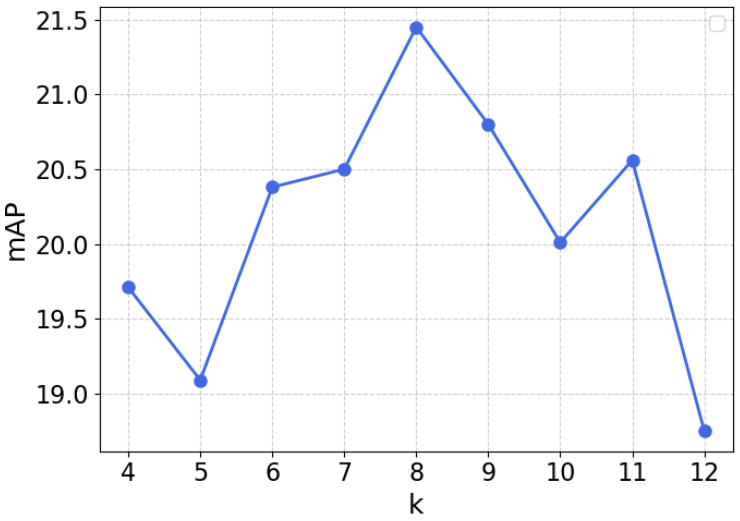}
    \caption{Ablation study of Top-K parameter \( k \) in PSSC on mAP performance.}
    \label{fig:k_ablation}
\end{figure}

\noindent\textbf{Ablation on PSSC Parameters.}
Our PSSC enhances emotion boundary localization by leveraging the semantic space across temporal sequences, as illustrated in Fig.~\ref{fig:pipeline}(c). The ablation study on the sample quantity in Eq.~\ref{eq:topk} is conducted based on \( K = \lfloor T / k \rfloor \), where \( T \) denotes the total number of frames. This design accounts for significant temporal variations across input videos.
As shown in Fig.~\ref{fig:k_ablation}, the hyperparameter \( k \) has a substantial impact on performance, with mAP showing a nonlinear trend. The best result (21.5\% mAP) occurs at \( k = 8 \), indicating a balance between emotional coverage and noise suppression.
A moderate sample count promotes the optimization of \(\mathcal{L}_{\text{sc}}\) by aggregating discriminative emotional patterns and capturing the spatial distribution of frame-level \( F_{\text{mix}} \). In contrast, \( k = 5 \) yields insufficient cues, while \( k = 12 \) introduces redundancy.

\noindent\textbf{Ablation on Loss Weights.}
\begin{figure}[tp]
    \centering
    \includegraphics[width=0.8\linewidth]{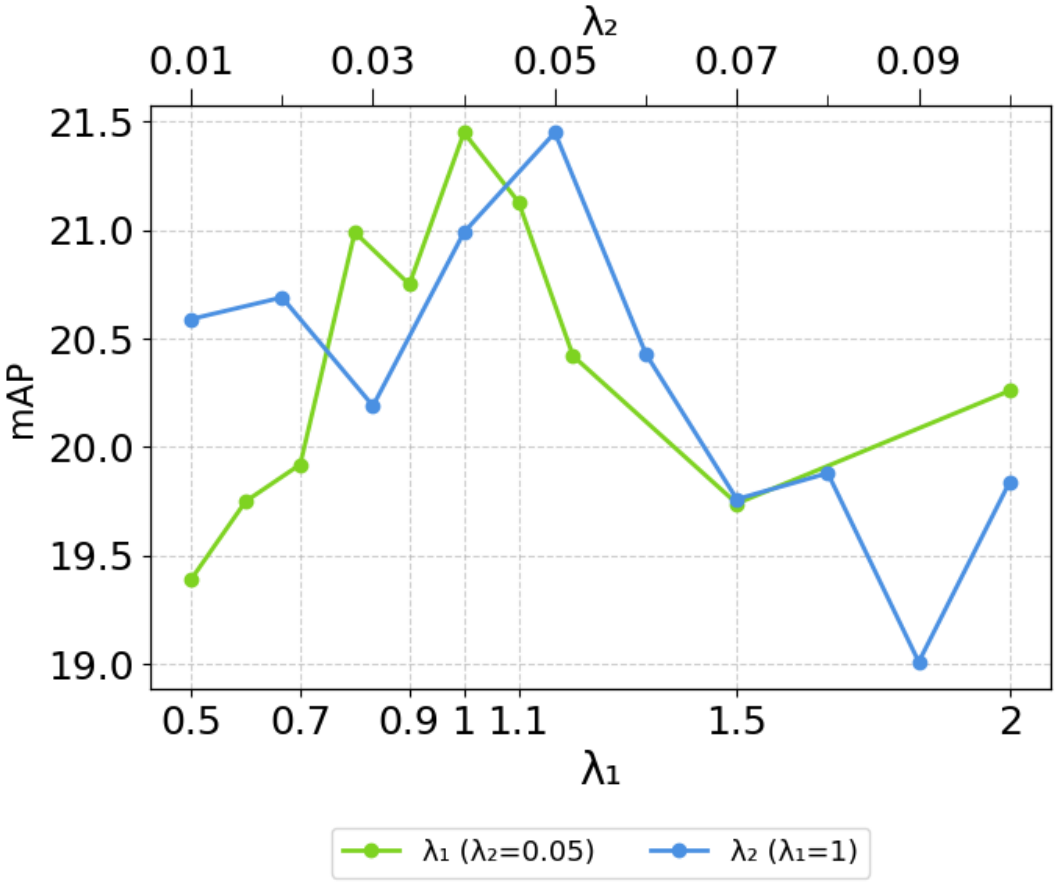}
    \caption{Ablation study of loss function weights $\lambda_1$ and $\lambda_2$ on mAP performance.}
    \label{fig:lambda_ablation}
\end{figure}
We explore the impact of loss weights on the experiment in terms of mAP. As shown in Fig.~\ref{fig:lambda_ablation}, the blue curve indicates the variation of \(\lambda_1\) when \(\lambda_2 = 0.05\), while the green curve shows the variation of \(\lambda_2\) when \(\lambda_1 = 1\). The mAP reaches its peak at \((\lambda_1, \lambda_2) = (1, 0.05)\), highlighting the importance of properly balancing the two loss terms. Specifically, \(\mathcal{L}_{\text{frame}}\) improves the accuracy of emotion localization, whereas \(\mathcal{L}_{\text{sc}}\) helps align frame-level emotion segments. A small \(\lambda_1\) results in insufficient boundary supervision, while an overly large \(\lambda_2\) may over-constrain the alignment, leading to blurred segment boundaries and loss of fine-grained emotional cues.

\section{Additional Results}
\begin{figure}
    \centering
    \includegraphics[width=0.99\linewidth]{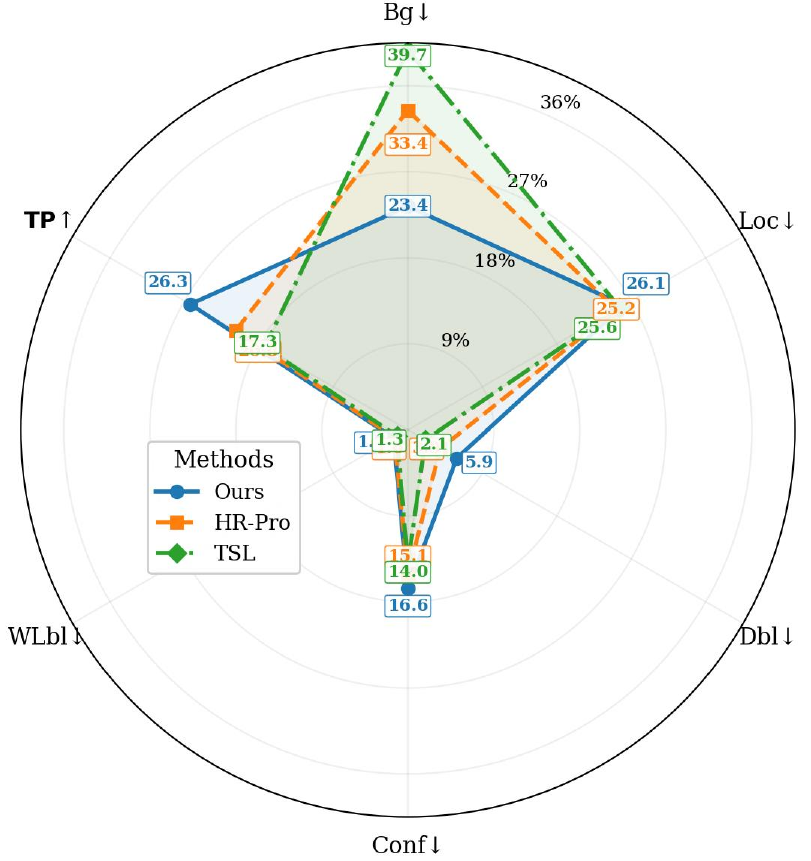}
    \caption{Proportion of proposal type for all prediction results of models on TSL300. 
Error categories: Bg (background), Loc (localization), Dbl (double detection), Conf (confusion), WLbl (wrong label), and TP (true positive).}
\label{fig:placeholder}
\end{figure}
\noindent\textbf{Error Analysis.}
To gain deeper insight, we follow the error diagnosis protocol of \cite{alwassel2018diagnosing} to break down the model outputs on the TSL300 dataset. As in Fig.~\ref{fig:placeholder}, a key observation is the substantial reduction in background errors, with a 10.0\% decrease compared to HR-Pro (33.4\% → 23.4\%) and a 16.3\% decrease compared to TSL (39.7\% → 23.4\%). This is accompanied by a notable increase in true positives (TP: 26.3\%). These results strongly suggest that our approach effectively reduces false positives arising from background confusion. The fact that localization error (Loc) remains the dominant error type also indicates potential for further improvement in overall detection accuracy (mAP), pointing to future work on refining temporal boundaries.

\noindent\textbf{PSSC Distance Metric Ablation.}
To assess the impact of different distance metrics on the identification of semantically relevant positive samples, we conduct an ablation study on the similarity function used in Eq.~(4) of the PSSC module.
Table~\ref{tab:distance_strategy_detail} reports the mAP results under multiple IoU thresholds. Among the compared metrics, \textbf{Cosine similarity (Ours)} achieves the highest average mAP of 21.45\%. In comparison, the L2 distance, the L1 distance and the dot product achieve lower scores of 20. 29\%, 20. 05\% and 21. 00\%, respectively. This corresponds to absolute improvements of 1.16\%, 1.40\%, and 0.45\%. Under low IoU thresholds (0.1, 0.15, 0.2), where spatial constraints are loose and semantic disentanglement is more difficult, Cosine still achieves the best performance, with mAP of 29. 31\%, 25. 47\%, and 22. 49\%.
These results demonstrate the effectiveness of cosine similarity in capturing semantic relationships between feature vectors.

\begin{table}[!t]
\centering
\renewcommand\arraystretch{0.9}
\caption{Performance comparison of different modality combinations on the TSL task. A: Audio, V: Visual, F: Local Face. Metrics are reported as mAP@IoU (\%) across various thresholds and average mAP.}
\resizebox{0.99\columnwidth}{!}{
\begin{tabular}{l|ccccc|c}
    \toprule
    \multirow{2}{*}{\textbf{Modality}} & \multicolumn{5}{c|}{\textbf{mAP@IoU (\%)}} & \textbf{Avg} \\
    & 0.1 & 0.15 & 0.2 & 0.25 & 0.3 &\textbf{mAP}\\ 
    \hline
    \rowcolor{gray!20}
    \multicolumn{7}{c}{\textit{Single Modality}} \\
    \midrule
    A (Audio)           & 18.98 & 15.47 & 12.96 & 9.75  & 7.09  & 12.85 \\
    V (Visual)          & 19.27 & 16.14 & 12.06 & 9.14  & 6.45  & 12.61 \\
    F (Local Face)      & 22.11 & 15.97 & 11.33 & 8.40  & 6.62  & 12.88 \\
    \hline
    \rowcolor{gray!20}
    \multicolumn{7}{c}{\textit{Bimodal Combinations}} \\
    \midrule
    A + F               & 29.41 & 24.01 & 18.70 & 13.97 & 9.95  & 19.21 \\
    A + V               & 25.91 & 23.70 & 20.08 & 15.85 & 11.53 & 19.41 \\
    V + F               & 26.20 & 22.48 & 17.41 & 14.53 & 11.30 & 18.38 \\
    \hline
    \rowcolor{gray!20}
    \multicolumn{7}{c}{\textit{Trimodal Combination}} \\
    \midrule
    A + V + F           & \textbf{29.31} & \textbf{25.47} & \textbf{22.49} & \textbf{16.76} & \textbf{13.24} & \textbf{21.45} \\
    \bottomrule
\end{tabular}}
\label{tab:modality_ablation}
\end{table}
\begin{table}[ht]
\centering
\small 
\caption{Ablation of feature distance metrics for Point-aware Sentiment Semantics Contrast modeling.}
\resizebox{0.99\columnwidth}{!}{
\begin{tabular}{c|ccccc|c}
\toprule
\renewcommand{\arraystretch}{0.87}
\multirow{2}{*}{\textbf{Method}} 
& \multicolumn{5}{c|}{\textbf{mAP@IoU (\%)}} 
& \multirow{1}{*}{\textbf{Avg}} \\
& 0.1 & 0.15 & 0.2 & 0.25 & 0.3 & \textbf{mAP}\\
\midrule
L2 & 28.06 & 24.40 & 20.98 & 16.49 & 11.55 & 20.29 \\
L1 & 26.53 & 23.66 & 20.70 & 16.36 & 12.97 & 20.05 \\
Dot& 27.23 & 24.90 & 21.21 & \textbf{17.42} & \textbf{14.22} & 21.00 \\
\textbf{Cosine(Ours)} & \textbf{29.31} & \textbf{25.47} & \textbf{22.49} & 16.76 & 13.24 & \textbf{21.45}\\
\bottomrule
\end{tabular}
}
\label{tab:distance_strategy_detail}
\end{table}

\noindent\textbf{Modality Combination Analysis.}
We conduct an ablation study to examine the impact of the modality information, as shown in Table~\ref{tab:modality_ablation}, which reports the performance of different modality combinations.
Among the unimodal inputs, the face embedding (F) achieves the best performance (mAP: 12.88\%), slightly outperforming the audio (A) and visual (V) features. This suggests that each modality provides complementary yet insufficient information when used in isolation.
Bimodal combinations such as A+V and A+F yield significantly improved results (mAP: 19.41\%, representing an increase of 6.56\%; and 19.21\%, with an increase of 6.36\%, respectively). These results indicate strong complementarity between audio and visual/facial cues. Meanwhile, the relatively lower performance of V+F (18.38\%) may be attributed to redundancy between holistic and local visual features. 
Trimodal fusion (A+V+F) further boosts the performance to 21.45\% mAP, demonstrating the advantage of integrating heterogeneous modalities.

\end{document}